\documentclass[10pt,twocolumn,letterpaper]{article}

\usepackage{3dv}
\usepackage{times}
\usepackage{epsfig}
\usepackage{graphicx}
\usepackage{amsmath}
\usepackage{amssymb}
\usepackage{caption}
\usepackage{subcaption}
\usepackage{multirow}
\usepackage{booktabs}       

\threedvfinalcopy 

\def\threedvPaperID{****} 

\ifthreedvfinal\fi
\begin{document}

\title{DDM-NET: End-to-end learning of keypoint feature Detection, Description and Matching for 3D localization}

\author{Xiangyu Xu$^1$ \quad Li Guan$^2$ \quad Enrique Dunn$^1$ \quad Haoxiang Li$^2$ \quad Gang Hua$^2$\\
$^1$Stevens Institute of Technology \quad $^2$Wormpex AI Research\\
}

\maketitle

\begin{abstract}
In this paper, we propose an end-to-end framework that jointly learns keypoint detection, descriptor representation and cross-frame matching for the task of image-based 3D localization. Prior art has tackled each of these components individually, purportedly aiming to alleviate  difficulties in effectively train a holistic network. We design a self-supervised image warping correspondence loss for both feature detection and matching, a weakly-supervised epipolar constraints loss on relative camera pose learning, and a directional matching scheme that detects key-point features in a source image and performs coarse-to-fine correspondence search on the target image. We leverage this framework to enforce cycle consistency in our matching module. In addition, we propose a new loss to robustly handle both definite inlier/outlier matches and less-certain matches. The integration of these learning mechanisms enables end-to-end training of a single network performing all three localization components. 
Bench-marking our approach on public data-sets, exemplifies how such an end-to-end framework is able to yield more accurate localization that out-performs both traditional methods as well as state-of-the-art weakly supervised methods.
\end{abstract}

\section{Introduction} \label{sec:intro}

Feature key point detection, descriptor design and feature matching are fundamental building blocks for visual localization. Traditional hand-crafted feature descriptors such as SIFT and their matching have been proved successful for many scenarios with rich textures and moderate imaging geometry. With the recent advances in deep learning, trained feature descriptors \cite{mishchuk2017working} \cite{simo2015discriminative} \cite{yi2016lift} and matching networks \cite{sarlin2020superglue} have shown promising results on some well-known benchmark data-sets \cite{li2018megadepth} \cite{dai2017scannet}. However, in order for the learned methods to perform on-par with the traditional methods in general scenes, it is critical to feed the networks high-quality labelled data in both quantity and variation \cite{luo2018geodesc} \cite{schonberger2017comparative}. For image pose localization task, the ground truth labels of corresponding feature sub-pixel  locations are tedious to obtain, since it fundamentally is not a friendly task designed for human.

To alleviate the dependency on strong supervision and costly human labeling, recent papers have proposed to use weak supervision on geometric constraints such as relative camera pose information on image pairs that can be obtained from state-of-the-art SfM pipelines \cite{schoenberger2016sfm}, \cite{Moulon2012}.
This has been proven to be successful at learning low-level feature detection \cite{li2022decoupling} and feature descriptors \cite{wang2020learning} separately, with the additional supervision of a traditional matching module behind it. 
It would be beneficial to have a pipeline that incorporate key point detection, feature descriptor learning and correspondence matching in an end-to-end fashion. 
The motivation is that training each component separately and individually may only result in modules that introduce systematic bias, and tops as good as the rest fixed modules can perform. Whereas by training all modules together, data knowledge can be communicated from matching stages all the way back to feature proposal and descriptor learning, thus realizing the full potential of the training data. Additionally, to train keypoint detection, feature descriptor and matching together would streamline fine-tuning on new data-sets as all stages shall be adapted jointly.

The main challenge lies again in lack of labelled data. With full-supervision, when feature locations and matching correspondences are provided, learning risks converging very quickly, over-fitting to the training data and compromising performance generalization. Conversely, when learning is based only weak supervision such as image pair poses, network architecture, losses, and training strategies need to be carefully designed to make sure the whole pipeline remains trainable.

In this paper, we propose a weakly supervised and self-supervised framework that jointly trains feature detection, descriptor extraction and correspondence computation in an end-to-end network. We deploy a directional matching backbone that detects keypoint features in our source image and performs coarse-to-fine correspondence search on our target image. Our detector is self-supervised by the weighted 8-point algorithm to learn each keypoints' contribution to recover relative poses based on image appearance. Similar to \cite{wang2020learning}, the only external supervision required for training is the pre-computed relative camera poses, which provide the epipolar constraints between image pairs. 

Our hierarchical correspondence search framework only enforces epipolar consistency on the finest image level for improved geometric accuracy, while relying on the appearance-based feature map content similarity to guide our coarse-to-fine matching and propagation. Such an approach promotes coarse levels to focus on yielding robust visual appearance encodings, while having finer levels focus on geometric accuracy.

In addition, we propose to add a self-supervision paradigm to regulate the matching training, similarly to the approach in \cite{detone2018superpoint}. After the feature extraction sub-network, randomly warped images with ground-truth homography matching correspondences are augmented for the matching sub-network. We show this is a vital step to the success of our training. Finally, since we do not rely on the ground truth matching supervision, we design new robust loss "outliers robust distance function" to handle potential outliers, inliers and non-overlapping regions during training.

We show improved performance on relative pose localization in both indoor and outdoor data-sets against state-of-the-art weakly supervised localization approaches \cite{wang2020learning}. Our general framework is depicted in Figure 1. 
The main contributions of our paper are as follows:
\begin{enumerate}
    \item{\bf Expanding the scope of learned low-level visual data association}. We propose the first end-to-end learned pipeline to jointly address the tasks of feature detection, description and matching.
    \item{\bf Directional coarse-to-fine matching module} We propose a matching module combining of a continuous coarse-to-fine layer to estimate feature correspondence based on multi-level images features.
    \item {\bf Learning from limited supervision signals}. We 
    rely exclusively on epipolar geometry  ground truth (weak-supervision) and image warping  strategy (self-supervision) to train the whole pipeline.
    \item{\bf A robust loss function for geometric matching error}. We propose a three-parameter combination of Huber and truncated loss that simultaneously bounds the contribution of gross outliers, while adaptively dampening the loss contribution of geometrically ambiguous correspondences.
\end{enumerate}
Benchmarking our approach on public data-sets,  exemplifies how joint training of all components in an end-to-end framework is able to yield more accurate localization performance over both traditional methods as well as state-of-the-art weakly supervised methods.

\section{Related Work} \label{sec:related}





\subsection{Learning based feature matching}
With the fast development of deep learning methods proposed for solving feature matching problem, many learning-based methods of feature detector, feature descriptor, feature matching and robust estimation are now competitive against the classical hand-craft methods such as SIFT \cite{lowe2004distinctive} and RANSAC \cite{fischler1981random}. Feature patch learning methods \cite{keller2018learning,luo2018geodesc,ono2018lf,simo2015discriminative,tian2017l2,yi2016lift} are designed
as the direct replacement of hand-crafted feature descriptors, which are designed for detecting repeatable keypoints from an image pair  and extracting features from local image patches for correspondence search.
Conversely, learned dense feature descriptors \cite{liu2019gift,revaud2019r2d2,lin2017feature,sun2021loftr,wang2020learning} defined over the whole image instead of only local visual information and their corresponding feature matching methods \cite{sarlin2020superglue,rocco2018neighbourhood,rocco2020efficient,li2020dual,zhou2021patch2pix}  estimate the best correspondences based on these dense features. Dense descriptor and matching methods are giving state-of-the-art results, at the cost of much more computational power comparing to sparse feature methods. Conversely, learning-based keypoint detectors \cite{detone2018superpoint,tian2020d2d,christiansen2019unsuperpoint,jau2020deep,kim2021self} have been proposed, as well as outlier rejection methods \cite{yi2018learning,zhang2019learning}.

\subsection{Weakly supervised feature correspondence learning}
One of the main challenges for learning-based geometric perception methods is the selection and integration of a suitable supervisory signal, due to the expensive nature of ground-truth labels. Methods such as \cite{sarlin2020superglue,sun2021loftr} are supervised by ground truth matches estimated from poses and depth (e.g. from SfM methods). However, training feature descriptor and matching networks without explicit ground-truth matching is important both conceptually and in practice. The matching method in \cite{zhou2021patch2pix} is only supervised by epipolar geometry by first matching patch-level then pixel-level correspondence. CAPS \cite{wang2020learning} uses cycle consistency with a coarse-to-fine matching architecture to successfully learn feature descriptor only from camera positions. The work in \cite{spencer2020same} learned a Seasonal Invariant descriptor with only rough cross-seasonal image alignment. Examples of weakly supervised systems  optimizing matching score include \cite{rocco2018neighbourhood,rocco2020efficient}. Multiple learning-based outlier rejection methods \cite{brachmann2019neural,yi2018learning,zhang2019learning,sun2020acne} are also weakly supervised based on epipolar geometry without ground-truth correspondence labels.

\subsection{Self-supervised geometric perception}
The self-supervised method \cite{yang2021self} proposed an architecture based on existing feature matching networks and traditional outlier rejection methods (e.g. RANSAC). Feature matching generated from hand-craft feature descriptor is used as initialization to iteratively estimate a geometric model and train a matching network. Keypoint detectors \cite{detone2018superpoint,christiansen2019unsuperpoint,tian2020d2d,kim2021self} have been weakly supervised by generating virtual labels through ground-truth homography transformations to enhance keypoint repeatability.

\begin{figure*}[t!] 
    \centering
	\includegraphics[scale = 0.6,trim={0cm 0cm 0cm 0cm},clip]{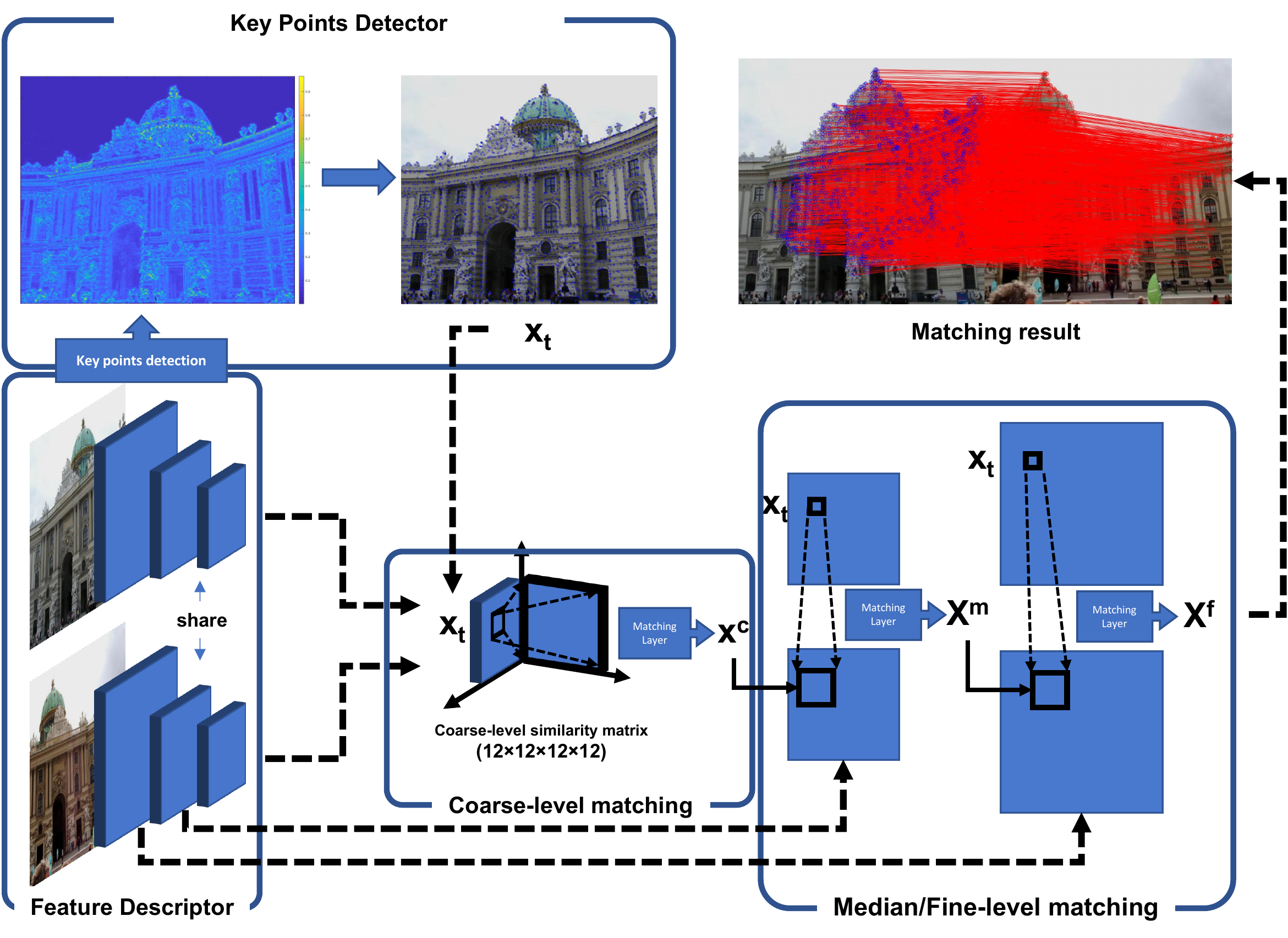}
	\caption{(a) \textbf{Feature descriptor} extracts multi-level feature map from images which is based on VGG\cite{balaban2015deep} architecture. (b) \textbf{Key point detection layer} estimates pixel-wise confidence on original image using a small CNN layer (c) \textbf{Coarse-to-fine matching} module gradually determines the feature correspondence location based on the feature similarity from target key points to a region of features in reference image and the coarse-to-fine correspondence is both continuous and differentiable, }
	\label{fig:network}
\end{figure*}

\begin{figure*}[t!] 
    \centering
	\includegraphics[scale = 0.5,trim={0cm 0cm 0cm 0cm},clip]{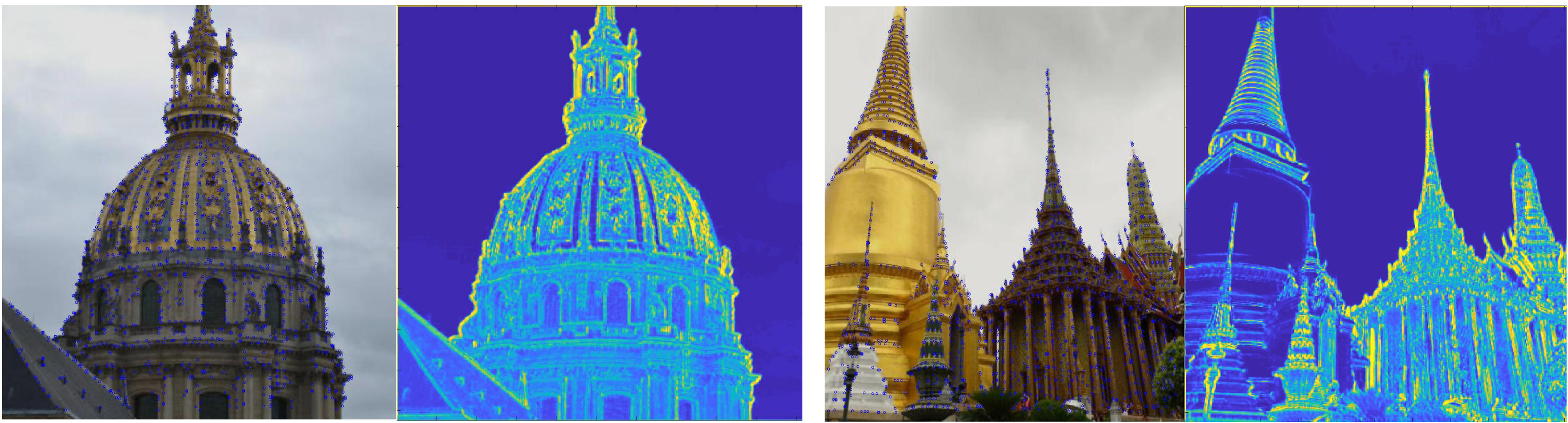}
	\caption{Examples of keypoint selection in blue (left) and its confidence map in yellow (right) . In an end-to-end manner, our method effectively learns which keypoints are potentially amenable to matching based on image local appearance. Such selection confidence maps are computed at full image resolution through a fully connected layer. Best viewed in color.}
	\label{fig:key points}
\end{figure*}

\section{Method} \label{sec:method}

Given a target image $I_t$ and reference image $I_r$ observing the same scene, we propose a method for end-to-end estimation of the dense local features, keypoint selection and feature correspondence between the two images by leveraging only the supervision of ground truth relative camera poses, which can be obtained from state-of-the-art SfM pipelines \cite{schoenberger2016sfm}, \cite{Moulon2012} given image sequence data-sets for training. During testing time, only the two images $I_t$ and $I_r$ are required, and the network computes the relative pose between them as the output. In Sec. \ref{Sec:feat_extract}, \ref{Sec:detect}, and \ref{Sec:matching}, each module of our network is introduced as feature extractor, detector to matching modules. Then, we talk about our coarse-to-fine matching architecture in Sec. \ref{Sec:coarse_to_fine}, followed by the loss formulations for training our network in the presence of outliers in Sec. \ref{Sec:loss}. Finally, we explain in detail how we bootstrap our  keypoint detector and fine-tune the matching in two steps in Sec. \ref{Sec:training}. An overview of our method is presented in Fig. \ref{fig:network}.

\subsection{Feature Extraction Layer}
\label{Sec:feat_extract}
We first use a convolutional architecture based on VGG \cite{balaban2015deep} to extract dense local features at multi-level resolution for both $I_t$ and $I_r$. As in Fig. \ref{fig:network}, we keep features $F^f$, $F^m$, and $F^c$, which are respectively: fine-level features at $1/2$ of original resolution, median-level features at $1/8$ of original resolution, and coarse-level features at $1/32$ of original resolution which is followed by adaptive average pooling to resize to $16 \times 16$ resolution. The weights of the network are shared between two images and the resolution of the features are the same for both images.

\subsection{Keypoint Detection Layer}
\label{Sec:detect}
In Fig. \ref{fig:network}, we process the target image with a small convolutional network followed by a  $Sigmoid$ to estimate a confidence map $C$ and constrain the values in the range from $0$ to $1$ at the same resolution as the original images for keypoints detection. Then, we apply non-maximum suppression to choose coordination $x_t$ with confidence larger than a threshold as the ``keypoints'' for matching. Different from the traditional keypoints detectors which extract key points from both images and require repeatability, our detector only estimates keypoints on the target image, striving to identify those image locations that would be easy to match in reference image. Some key points detection examples are showed is Fig. \ref{fig:key points}
\subsection{Matching Layer}
\label{Sec:matching}
The role of our matching layer is to find the best 2D correspondence of a target image keypoint inside a specific region of the reference image, based on a pairwise feature similarity measure. Given the dense feature and keypoint's position $x_t^k \in R^2$ of target image and the feature patch $P_r$ of a specific region in reference image of size $(W_r \times W_r)$, we first crop a feature patch $P_t$ of size $(W_t \times W_t)$ at $x_t^k$ in the target image. Then, we calculate the pairwise similarity among patches by dot product followed by softmax between $P_r$ and $P_t$ and reshape it to a similarity matrix of size $(W_t^2 \times W_r \times W_r)$. Finally, we process the similarity matrix with a small convolutional network followed by a fully connected layer and a $Sigmoid$ to normalize it in the range from $0$ to $1$. The output coordinates $x_r^k \in R^2$ represents  the relative location in the reference image region that corresponds to the target image key point $x_t^k$, where $x_r^k =(0,0)$ corresponds to  the upper-left corner.

\subsection{Coarse-to-Fine matching module}
\label{Sec:coarse_to_fine}
The main advantages of our coarse-to-fine matching module are (1) differentiability with respect to network parameters; (2) continuity from coarse to fine matching; (3) robustness to matching feature with different scales.


\subsubsection{Coarse-level matching}

In the coarse-level matching layer, for a keypoint location in the target image $x_t^k$, we
extract a local window centered on it, of size $W_t$.  From this window we compute coarse-level similarity matrix $S_c^k$ which is computed with respect to all of the ($ W_c \times W_c $) locations in the coarse reference image. 
We perform exhaustive evaluation of such similarity values, yielding a tensor $S_c$ of dimensions $(W_t \times W_t \times W_c \times W_c)$. Finally, the coarse level matching coordinates $x_r^{c,k}$ for a key point $x_t^k$ is estimated through the matching layer based on our pairwise similarity matrix $S_c^k$.


\begin{figure*}[t!] 
    \centering
	\includegraphics[scale = 0.4,trim={0cm 0cm 0cm 0cm},clip]{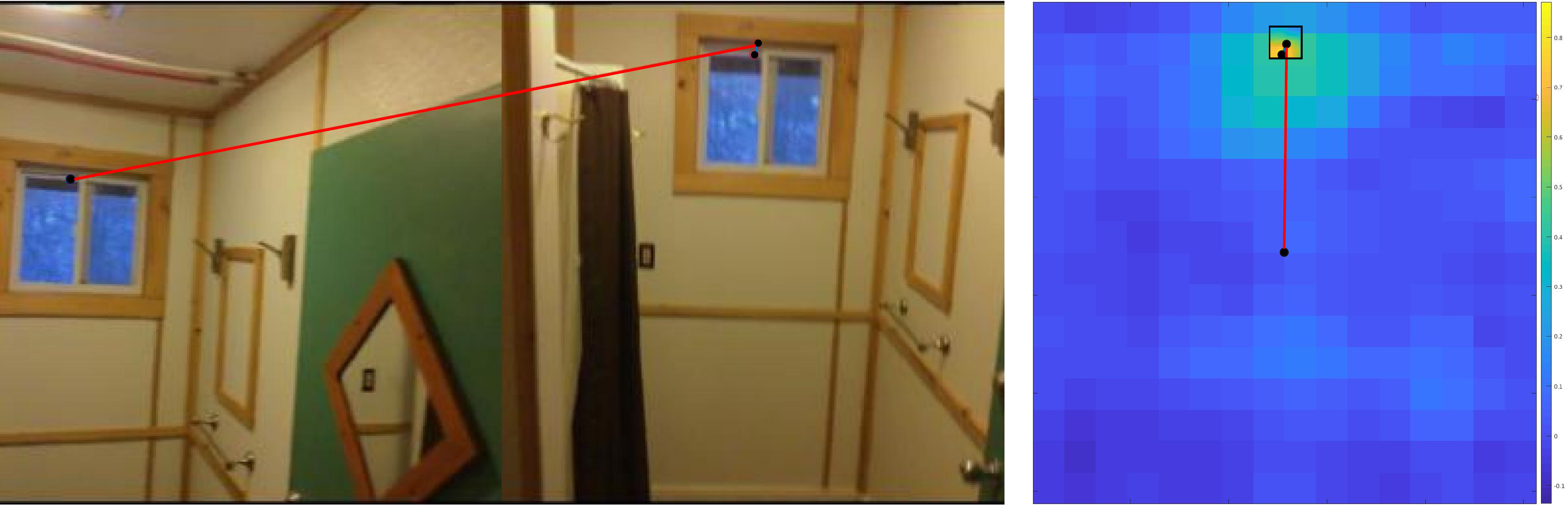}
	\caption{Directional feature correspondence search across different resolutions/scope. For a given feature location,  matching layers compute the  correspondence location based on localized coarse-to-fine similarity maps  (a) At left: Given the limited spatial extent/scope, \textbf{coarse-level matching} is computed from the whole reference image. (b) Once coarse matching is attained at lower resolutions, \textbf{fine-level matching} upscales said matching location, defines a local window centered around it and estimate a local refinement offset vector.}
	\label{fig:c2f_matching}
\end{figure*}

\subsubsection{Median-level and Fine-level matching}
For example median-level matching, for a keypoint $x_t^k$,  feature patches $P_t^{m,k}$ and $P_r^{m,k}$ are extracted from the median-level features $F_r^m$ and $F_t^m$ separately at the center of $x_t^k$, with the window size of $W_t$, and $x_r^{c,k}$ with the window size of $W_r$. Then, we compute the similarity matrix $S_m^k$ based on the feature patches. After going through our matching layer, median-level correspondence coordinates $x_r^m$ are estimated inside the region $(W_r \times W_r)$ at $x_r^{c,k}$ and the final correspondence coordinates after both the coarse and median-level matching is simply the vector sum $ x_r = x_r^c + x_r^m$. Finally, we do the same at fine level feature and fine-level matching will be $ x_r = x_r^{c,k} + x_r^{m,k} + x_r^{f,k}$  as shown in Fig. \ref{fig:c2f_matching}. 

As seen in Fig. \ref{fig:c2f_matching}, the red line in the right figure is the coarse match result which finds the correspondence around the right match, the left figure presents the coarse level similarity heatmap. Then, by searching within the coarse matching region, a more accurate correspondence location (blue line) can be estimated. Fine-level similarity heatmaps around the coarse matching coordinate are shown in the black point in the target image on the left.

\subsubsection{Continuity of our feature patch representation}
In order to make the coarse-to-fine matching process, continuous and differentiable, each feature in the patch is estimated through bi-linear interpolation among the corresponding neighboring features. 

\subsection{Loss Formulations}
\label{Sec:loss}
\subsubsection{Epipolar Loss}
Given a pair of matching coordinates $(x_t, x_r)$, intrinsic matrices $(K_r, K_t)$ and an essential matrix $E$ computed from the ground-truth relative pose (provided with the training data from SfM or a similar process). We penalize the matching based on symmetric epipolar distance per the equations below.
\begin{equation}
    d_e(p_t,p_r,E) = \frac{p_rEp_t}{\sqrt{(Ep_t)^2_{[1]}+(Ep_r)^2_{[2]}}}
\end{equation}
\begin{equation}
    L_e(p_t,p_r,E) = d_e(p_t,p_r,E) + d_e(p_r,p_t,E^T)
    \label{Eq: Ep Loss}
\end{equation}
where $p_t = K_t^{-1}[x_t,1]^T$ and $p_r = K_r^{-1}[x_r,1]^T$ are the normalized coordinates.

\subsubsection{Reverse Matching Consistency Loss}
The motivation for this loss formulation is that for a true matching $x_t \rightarrow x_r$ from the $I_t$ to $I_r$, and the reverse matching from $x_r \rightarrow x_c$ from the $I_t$ to $I_r$, the distance from $x_c$ to $x_t$ in $I_t$ should be as small as possible. Hence, we penalize the reverse matching consistency loss as following,
\begin{equation}
    L_{cy}(x_t,I_t,I_r) = d_{cy}(g_{I_r \rightarrow I_t }(g_{I_t \rightarrow I_r }(x_t)) - x_t)
    \label{Eq:cy loss}
\end{equation}
where $x_r = g_{I_t \rightarrow I_r }(x_t)$ computing the matching coordinates from the keypoint $x_t$ on the image $I_t$ to the image $I_r$, $d_{cy}(\cdot)$ is a distance function. 

\subsubsection{Outlier-Robust Loss Function}
Our correspondences are weakly supervised by only ground-truth relative camera poses without any matching labels, so we need to avoid training the matching scheme according to potential outliers. We propose an outlier-robust loss function which is a combination of Huber Loss and truncated loss as following,
\begin{equation}
d(a)=
\begin{cases}
\frac{1}{2}a^2 & \text{for $|a| \leq  \delta_1$ }\\
\delta_1(|a|-\frac{1}{2}\delta_1) & \text{for $\delta_1 < |a| \leq \delta_2$}\\
\delta_1(\delta_2-\frac{1}{2}\delta_1) & \text{for $|a| > \delta_2$}
\end{cases}
\label{Eq:robust loss}
\end{equation}
where $\delta_1$ and $\delta_2$ are two positive value and $\delta_2>\delta_1$. It is applied for both $d_e$ and $d_{cy}$. See in Fig. \ref{fig:weakly-supervised}

\subsubsection{Keypoint Classification Confidence Loss}
Given the matching confidence $C_t$ of a set of key points $x_t$ from target images and their labels $y_t$, where $y_t^k = [0, 1]$ and $y_t^k = 1$ denotes the inlier correspondence, we define our key point classification confidence loss as,
\begin{equation}
    L_c(c_t,y_t) = H(c_t, y_t)
\end{equation}
where $S(\cdot)$ is the logistic Sigmoid function and $H(\cdot)$ is the binary cross entropy.

\subsubsection{Weighted Matching Loss}
For a given a pair of image $(I_t, I_r)$, a set of keypoints and the corresponding confidence $c_t$ on target image $I_r$, and having knowledge of the ground truth matching coordinates $x_r^{gt}$ on reference image $I_r$, we  formulate the weighted matching loss function as follows:
\begin{equation}
    L_m(I^{P},x_t,x_r^{gt},c_t) = (c_t/\bar{c_t}))||g_{I_t \rightarrow I_r }(x_t) - x_r^{gt}||^2
\end{equation}
where $\bar{c_t}$ is the mean of $c_t$, $I^P$ is a image pair $(I_t, I_r)$

\begin{figure*}[t!] 
    \begin{subfigure}[h!] {0.5\textwidth}
        \centering
    	\includegraphics[scale = 0.34,trim={0cm 0cm 0cm 0cm},clip]{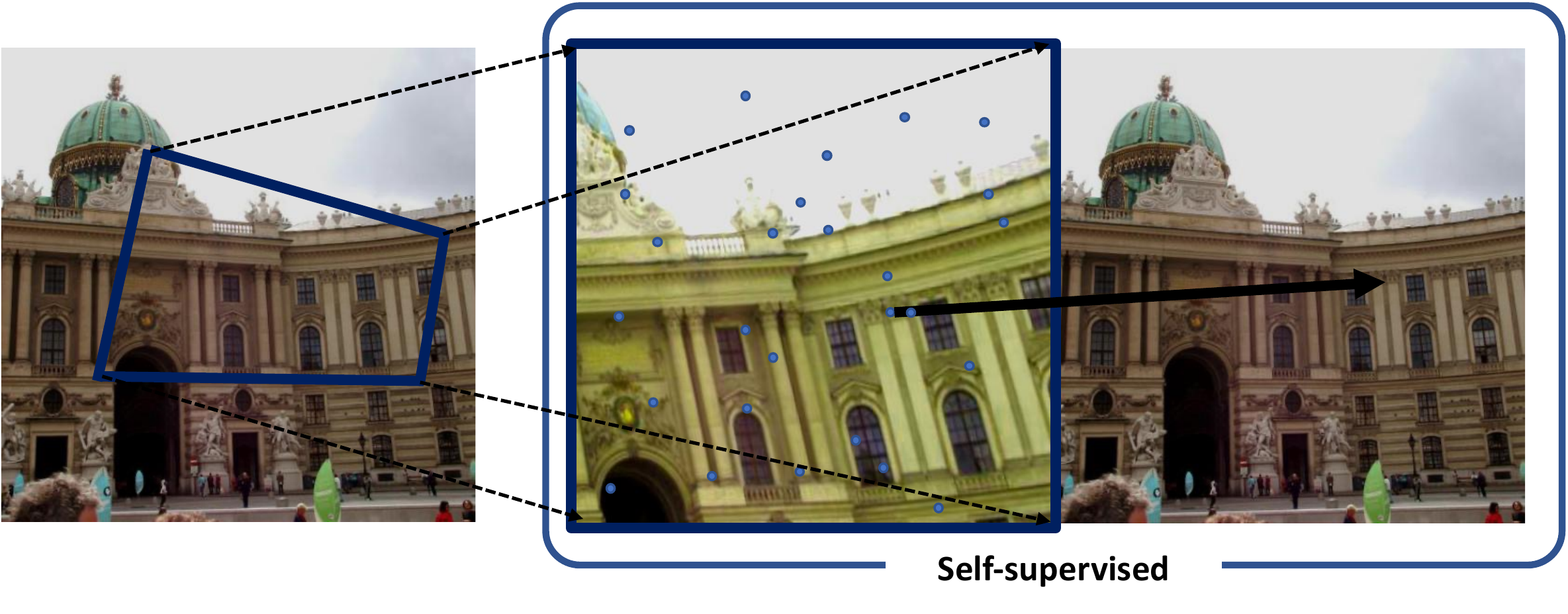}
    	\caption{Self-supervision}
    	\label{fig:self-supervised}
    \end{subfigure}
    \begin{subfigure}[h!] {0.5\textwidth}
        \centering
    	\includegraphics[scale = 0.26,trim={0cm 0cm 0cm 0cm},clip]{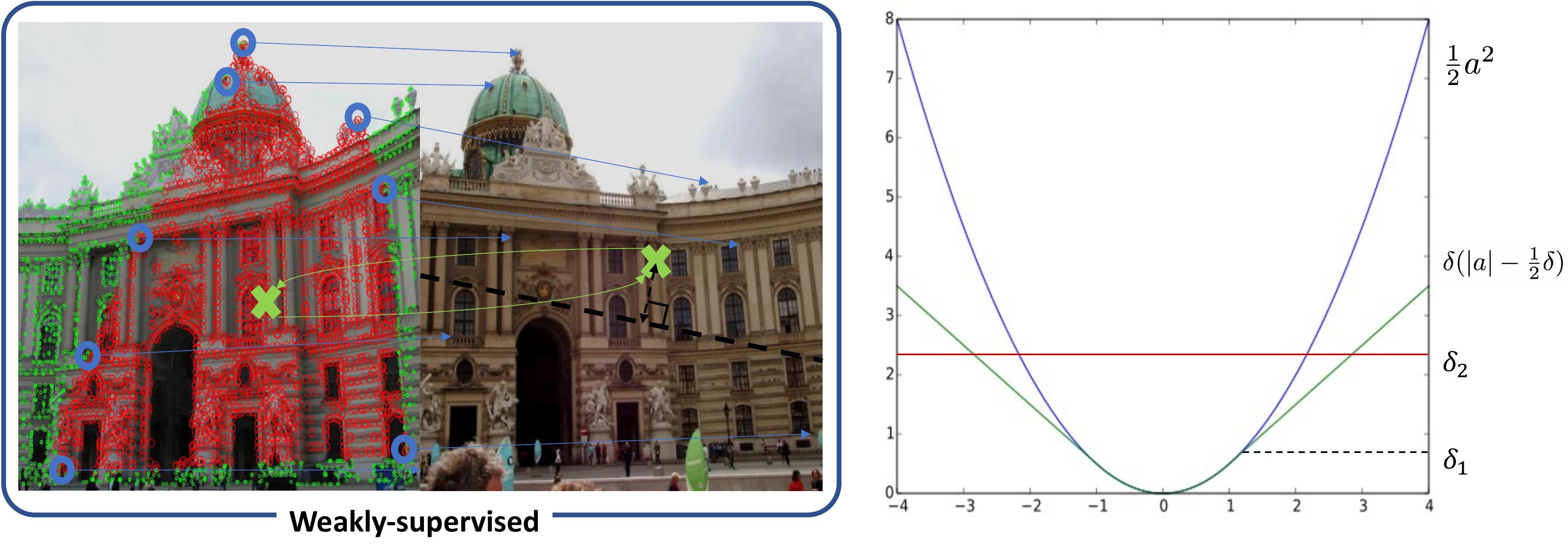}
    	\caption{Weakly-supervision}
    	\label{fig:weakly-supervised}
    \end{subfigure}
    \label{fig:training}
    
    \caption{Different supervision used by our approach. (a) \textbf{Self-supervision step} generates an image pair based on a random homography transformation of a single image. Such deterministic augmentation ensures  ground truth feature correspondences are available. (b) \textbf{Weakly-supervision step} first determines a few high confidence inlier matches based on the self-supervision training result and then determines additional candidates by finding the keypoints inside the convex hull they create. Finally, we penalize the Epipolar loss and reverse matching consistency loss with the help of our  robust loss function.}
\end{figure*}

\subsection{Training Strategy}
\label{Sec:training}
We propose two training strategies. (1) self-supervised strategy which only requires single-image warping. (2) Weakly-supervised strategy which is supervised by real pair image information and ground-truth relative poses.

\subsubsection{Self-supervised and Keypoint Training}
In this step, we are trying to train our network based on single-image warping. As in Fig. \ref{fig:self-supervised}, we first generate virtual image pairs by random homography transformations. Under this condition, for the any key point $X_r$ at target images, we can compute the ground truth matching coordinate $x_r^{gt}$ and use it for supervision. 

The main goal in this step is to train the parameters of the keypoint detection layer, so we randomly sample key points on the target images to learn the matching confidence based on its local appearance. The loss function used in this step:
\begin{equation}
    L_s = L_m + \lambda_1L_c
\end{equation}

\subsubsection{Weakly-supervised Matching Step}
After the keypoint detection layer is well-trained, we freeze the parameters of the detector layer, and sample the key points using it with non-maximum suppression instead of random sampling as in Fig. \ref{fig:weakly-supervised}. In this step of training, we add real image pairs generated by an image retrieval method.
For the real image pairs datasets, we only use the ground truth relative pose without actual matching label. The challenge is that we don't have the information either of exact overlapping area or point-wise feature correspondence  between each image pairs. To solve those challenges, we proceed as follows.

 First, as in Fig. \ref{fig:weakly-supervised}, based on the training result on the self-supervised step, we compute the keypoints and their feature correspondence as candidates for training. Then, we compute their reverse matching and estimate Epipolar loss and reverse matching consistency loss as in Eq. \ref{Eq: Ep Loss} and Eq. \ref{Eq:cy loss}. We keep the correspondence candidates as inliers if their losses are smaller than a threshold. Then, we estimate the convex hull based on the inlier key points and include all the key points inside the convex hull as candidates which will participate in the training.

Without ground truth feature correspondence, we penalize matching error by combination of the Epipolar loss and the reverse matching consistency loss. Then, the loss function used are as following,
\begin{equation}
    L_w = L_m + \lambda_3(L_e + L_{cy})
\end{equation}
There are still outliers that don't actually have a correspondence in the reference image, which could weaken the training result and reduce localization performance. We implement both Epipolar loss and reverse matching consistency loss with our robust loss function in Eq. \ref{Eq:robust loss} to reduce the effect of outliers.

\begin{figure*}[t!] 
    \centering
	\includegraphics[scale = 0.9,trim={0cm 0cm 0cm 0cm},clip]{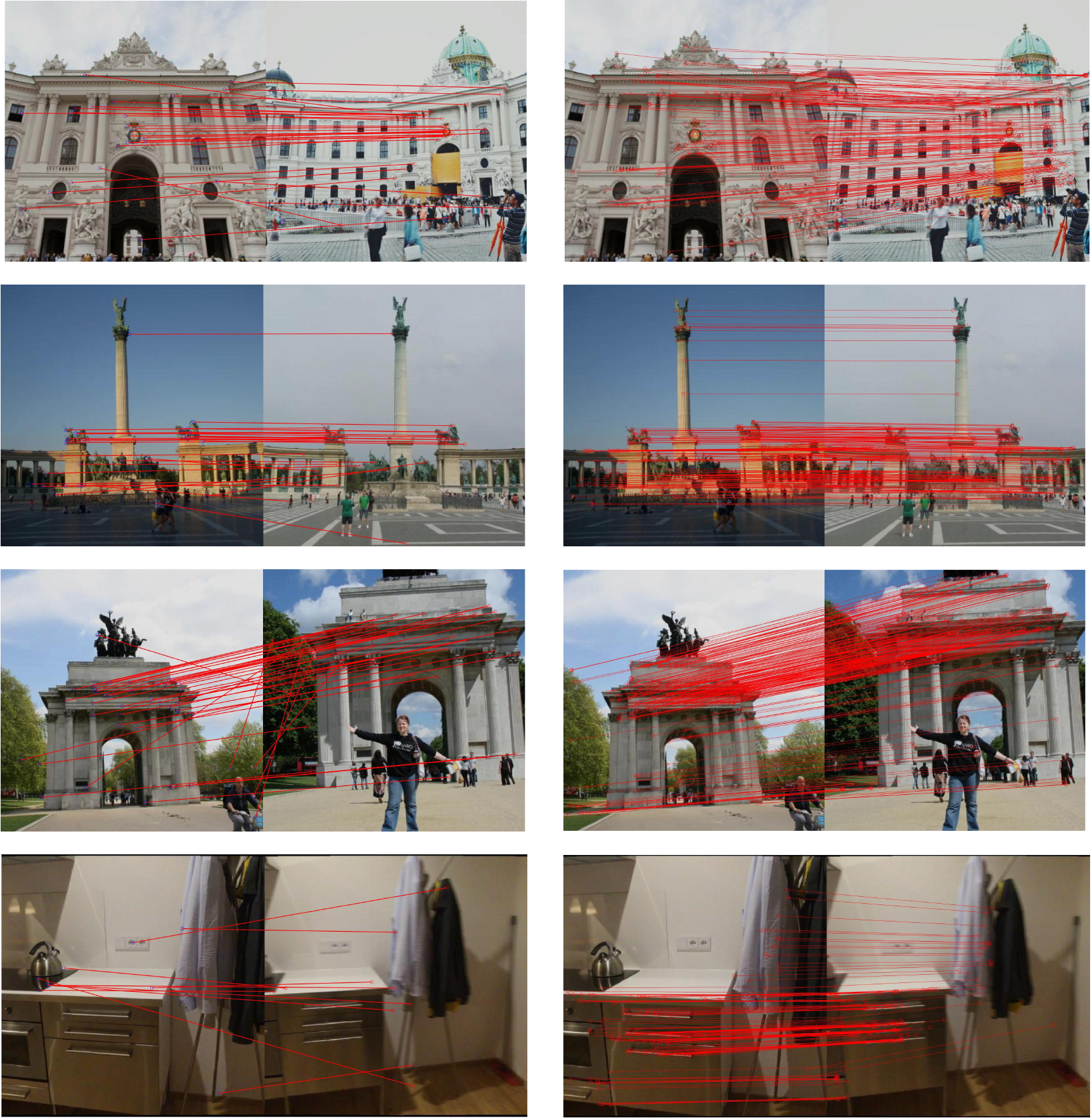}
	\caption{Qualitative comparison vs. SIFT putative matching followed by RANSAC. (a) Image-pair results on left column depict  SIFT-based matching. (b) Image-pair results on right column are the corresponding result for DDM.}
	\label{fig:examples}
\end{figure*}

\begin{table*}[t]
    \begin{center}
    \footnotesize
    \setlength{\tabcolsep}{3.3pt}
        \caption{\small \textbf{Rotation and translation accuracy on both ScanNet~\cite{dai2017scannet} and MegaDepth~\cite{li2018megadepth} datasets.} The accuracy of estimated rotations and translations are showed in the table below. We calculate the percentage of pairs with relative pose error under a certain threshold ($5^\circ$ for ScanNet and $10^\circ$ for MegaDepth).  $d_{\text{frame}}$ represents the interval between frames. Larger frame intervals imply harder pairs for matching. 
    }
     \label{tab:feature matching}
    \resizebox{\linewidth}{!}{%
    \begin{tabular}{lcccccc}
    \toprule
    \multirow{2}*{
    \textbf{Methods}
    } &\multicolumn{3}{c}{\textbf{Accuracy on ScanNet [\%]}} & \multicolumn{3}{c}{\textbf{Accuracy on MegaDepth [\%]}}\\ %
    \cmidrule(l){2-4} \cmidrule(l){5-7}
    & \textit{$d_{\text{frame}}$ = 10} & \textit{$d_{\text{frame}}$ = 30} & \textit{$d_{\text{frame}}$ = 60} & \textit{easy} & \textit{moderate} & \textit{hard} \\ %
    \midrule
    SIFT~\cite{lowe2004distinctive}& 
    91.0 / 14.1 & 65.1 / 15.6 & 41.4 / 11.9&
    58.9 / 20.2 & 26.9 / 11.8 & 13.6 / 9.6\\
    SIFT w/ ratio test~\cite{lowe2004distinctive}& 
    91.2 / 15.9 & 67.1 / 19.8 & 44.3 / 15.9&
    63.9 / 25.6 & 36.5 / 17.0 & 20.8 / 13.2\\
    SuperPoint~\cite{detone2018superpoint}& 
    94.4 / 17.5 & 75.9 / 26.3 & 53.4 / 22.1&
    67.2 / 27.1 & 38.7 / 18.8 & 24.5 / 14.1\\
    HardNet~\cite{mishchuk2017working} & 
    95.8 / 18.2 & 79.0 / 24.7 & 55.6 / 21.8 &
    66.3 / 26.7 & 39.3 / 18.8 & 22.5 / 12.3 \\
    LF-Net~\cite{ono2018lf}&
    93.6 / 17.4 & 76.0 / 22.4 & 49.9 / 18.0 &
    52.3 / 18.6 & 25.5 / 13.2 & 15.4 / 11.1 \\
    D2-Net~\cite{dusmanu2019d2} & 
    91.6 / 13.3 & 68.4 / 19.5 & 42.0 / 14.6&
    61.8 / 23.6 & 35.2 / 19.2 & 19.1 / 12.2 \\
    ContextDesc~\cite{luo2019contextdesc} &
    91.5 / 16.3 & 73.8 / 21.8 & 51.4 / 18.5&
    68.9 / 27.1 & 43.1 / 21.5 & 27.5 / 14.1 \\
    \midrule
    CAPS w/ SIFT kp. &
    92.3 / 16.3 & 74.8 / 22.5 & 50.8 / 20.9 &
    70.0 / 30.5 & 50.2 / 24.8 & 36.8 / 16.1\\
    CAPS w/ SuperPoint kp. &
    96.1 / 17.1 & 79.5 / 27.2 & \textbf{59.3} / \textbf{26.1} & 
    72.9 / 30.5 & 53.5 / 27.9 & \textbf{38.1} / 19.2 \\
    \midrule
    DDM-S &
    95.3 / 16.9 & 75.5 / 24.8 & 51.8 / 22.1 & 
    77.1 / 43.2 & 53.7 / 40.9 & 33.9 / 32.0 \\
    DDM-W &
    \textbf{96.2} / \textbf{19.1} & \textbf{80.0} / \textbf{27.6} & 53.8 / 22.1 &
    \textbf{80.9} / \textbf{44.5} & \textbf{57.9} / \textbf{46.8} & 35.7 / \textbf{38.4}\\
    \bottomrule
    \end{tabular}}
    \end{center}
    \vspace{-1em}
\end{table*}

\begin{table}[t]
    \begin{center}
    \footnotesize
    \setlength{\tabcolsep}{3.3pt} 
    \caption{\small Study at coarse(c) and median(m) matching result.
}
     \label{tab:ablation study}
    \resizebox{\linewidth}{!}{%
    \begin{tabular}{lcccccc}
    \toprule
    \multirow{2}*{
    \textbf{Methods}
    } & \multicolumn{3}{c}{\textbf{Accuracy on MegaDepth [\%]}}\\ 
    \cmidrule(l){2-4} \cmidrule(l){5-7}
    & \textit{easy} & \textit{moderate} & \textit{hard} \\ 
    \midrule
    DDM-W-c &
    49.3 / 9.80 & 18.3 / 10.1 & 8.53 / 4.52\\
    DDM-W-m &
    73.4 / 36.5 & 49.7 / 34.2 & 28.7 / 26.3\\
    \midrule
    DDM-W &
    \textbf{80.9} / \textbf{44.5} & \textbf{57.9} / \textbf{46.8} & \textbf{35.7} / \textbf{38.4}\\
    \bottomrule
    \end{tabular}}
    \end{center}
    \vspace{-1em}
\end{table}
\section{Experiments} \label{sec:exp}
We evaluate our method on both indoor data ScanNet \cite{dai2017scannet} and outdoor data MegaDepth \cite{li2018megadepth} for the task of estimating relative poses. Then, an ablation study quantifies the impact of each components of our method. 

\subsection{Relative poses estimation on MegaDepth}
We evaluate our method on the outdoor dataset MegaDepth \cite{li2018megadepth}, which has around 100 million image pairs and we sample 1 million from it for training. The testing image set are generated the same as in \cite{wang2020learning}, according to relative rotation angle: $easy ([0◦, 15◦])$, $moderate ([15◦, 30◦])$ and hard $([30◦, 60◦])$. See the results in Tab. \ref{tab:feature matching}.

\subsection{Relative poses estimation on ScanNet}
We use ScanNet \cite{dai2017scannet} as the indoor scene to show the performance of our method on relative pose estimation. The datasets includes about 2 billion image pairs. We only train on MegaDepth datasets and test on ScanNet to validate the generalization of our method. For testing, we use  generating images pairs as described in LF-Net \cite{ono2018lf}, which randomly samples image pairs at three different frame intervals, 10, 30, and 60. See the results in Tab. \ref{tab:feature matching}.

\subsection{Experiment setup}
We train our network on 8 GTX 1080 ti GPUs with batch size of 64. We resize the image as $512 \times 512 $ resolution for training and $800 \times 800 $ for testing. Based on our matching result, we estimate the essential matrix with OpenCV built-in RANSAC, then decompose the Essential matrix into rotation and translation. We report the accuracy of both translation and rotation in angles with metrics of percentage of frames under a threshold which is $5 \deg$ for ScanNet and $10 \deg$ for MegaDepth.

\subsection{Results}
In Tab. \ref{tab:feature matching}, we call our method as DDM, because we have a whole network pipeline for feature \textbf{D}etection, \textbf{D}escription, and \textbf{M}atching. We show the both result of self-supervised version (DDM-S) and weakly-supervised version (DDM-W) and compare with SOTA methods in the table. We can see that our method improve on both easy and moderate cases for MegaDepth datasets, competitive result on hard case, and significant boost on translation. Most important point is that our self-supervised version cam compete supervised and weakly-supervised result which is only trained on single images. Similar conclusion can be made in the ScanNet datasets result. Also see result in Fig. \ref{fig:examples}

\subsection{Ablation study}

As in Tab. \ref{tab:feature matching}, we did an experiment to study the accuracy of different level of matching in our method. In the tab \ref{tab:ablation study}, we can clearly see how finer level of matching improves the relative pose estimation on  the MegaDepth dataset.

\section{Conclusion} \label{sec:con}

In this paper, we propose a new 3D localization framework that trains in an end-to-end  manner the three major 3D localization components: feature detection, descriptor extraction and correspondence matching, with weak supervision only on camera poses. To guarantee effective training, we redesign the hierarchical correspondence search framework, which only enforces epipolar consistency on the finest image level for improved geometric accuracy, while relying on the appearance-based feature map content similarity to guide our coarse-to-fine matching and propagation. We adopt a relatively simple but surprisingly effective self-supervised image warping
 augmentation to foster accurate pixel-to-pixel correspondence for training our matching module.  We propose a robust loss to handle potential matching outliers. Experiments showed that our method outperforms state-of-the-art weak supervision approaches on both indoor and outdoor localization data-sets. The improvement indicates that the three components of the localization tasks are tightly coupled together and are able to help one another when trained together. Future work includes to evaluate the approach performance under different appearance conditions. We also like to explore the possibility to replace the RANSAC module for pose computation after matching to make a full visual localization network.

{\small
\bibliographystyle{ieee_fullname}
\bibliography{egbib}
}
\end{document}